\documentclass[letterpaper, 10 pt, conference]{ieeeconf}  % Comment this line out if you need a4paper

\IEEEoverridecommandlockouts                              % This command is only needed if 
\usepackage[utf8]{inputenc}
\usepackage{amsmath,amssymb,amsfonts}
\usepackage{algorithmic}
\usepackage{graphicx}
\usepackage{textcomp}
\usepackage[dvipsnames]{xcolor}
\usepackage{subfigure}
\usepackage[colorlinks=true,linkcolor=red, anchorcolor=green,citecolor=red, linkbordercolor=green]{hyperref}
\usepackage[noadjust]{cite}
\usepackage[capitalize]{cleveref}

\let\labelindent\relax
\usepackage{enumitem}

\usepackage{tikz}
\newcommand*\circled[1]{\tikz[baseline=(char.base)]{
		\node[shape=circle,draw,inner sep=0.5pt] (char) {#1};}}
		
\DeclareRobustCommand{\legendsquare}[1]{
  \textcolor{#1}{\rule{2ex}{2ex}}
}

\def\BibTeX{{\rm B\kern-.05em{\sc i\kern-.025em b}\kern-.08em
		T\kern-.1667em\lower.7ex\hbox{E}\kern-.125emX}}

\PassOptionsToPackage{pagebackref=true}{hyperref}
\AtBeginDocument{\hypersetup{pdfborder={0 0 1}}}

\title{\LARGE \bf Mobile Mapping Mesh Change Detection and Update}
\author{Teng Wu$^{1}$, Bruno Vallet$^{1}$ and Cédric Demonceaux$^{2}$ 
\thanks{This research has been funded by ANR (Agence Nationale de la Recherche-France) Project pLaTINUM (ANR-15-CE23-0010).}	
\thanks{$^{1}$ LASTIG, Univ Gustave Eiffel, ENSG, IGN, F-94160 Saint-Mande, France {\tt\small firstname.lastname@ign.fr}}
\thanks{$^{2}$ VIBOT ERL CNRS 6000, ImViA, Université Bourgogne Franche-Comté, France {\tt\small cedric.demonceaux@u-bourgogne.fr}}
}

\begin{document}	
	
\maketitle
\thispagestyle{empty}
\pagestyle{empty}

\begin{abstract}

Mobile mapping, in particular, Mobile Lidar Scanning (MLS) is increasingly widespread to monitor and map urban scenes at city scale with unprecedented resolution and accuracy. The resulting point cloud sampling of the scene geometry can be meshed in order to create a continuous representation for different applications: visualization, simulation, navigation, etc. Because of the highly dynamic nature of these urban scenes, long term mapping should rely on frequent map updates. A trivial solution is to simply replace old data with newer data each time a new acquisition is made. However it has two drawbacks: 1) the old data may be of higher quality (resolution, precision) than the new and 2) the coverage of the scene might be different in various acquisitions, including varying occlusions.
In this paper, we propose a fully automatic pipeline to address these two issues by formulating the problem of merging meshes with different quality, coverage and acquisition time. Our method is based on a combined distance and visibility based change detection, a time series analysis to assess the sustainability of changes, a mesh mosaicking based on a global boolean optimization and finally a stitching of the resulting mesh pieces boundaries with triangle strips. Finally, our method is demonstrated on Robotcar and Stereopolis datasets.
\end{abstract}

%\begin{IEEEkeywords}
%	change detection, seam line optimization, mesh processing, long term mapping
%\end{IEEEkeywords}

% for submit 
% first key word : mapping 
% second key word : Object Detection, Segmentation and Categorization 
% third key word : Range Sensing

% ICRA
% Mapping; Object Detection, Segmentation and Categorization; Recognition

% a work flow
%1 multi sessions + icp resgistration
%2 change detection
%3 seam line optimization
%4 seam line stitch
\section{INTRODUCTION}
Benefiting from the development of 2D and 3D cameras and ranger devices such as RGB-D cameras and laser scanners, 3D point cloud of both indoor and outdoor environments can be easily and efficiently acquired \cite{Geiger2013IJRR, huitl2012tumindoor} and processed to generate 3D maps which have many applications such as visualization, simulation and (autonomous) navigation. In dynamic environments such as cities, changes might happen at very different time scales: instantaneous (pedestrians, moving vehicles), hours (parked vehicles, garbage cans), days (temporary changes such as markets, road works,...), years (structural changes). We consider that a map should only contain sustainable objects/geometries, defined in this paper as having an expected lifespan greater than some application specific sustainability threshold $T_s$ (for mapping purposes we consider $T_s=1$ month to be a good threshold between temporary and structural changes).

\begin{figure*}[t]
	\centering
	\includegraphics[width=\linewidth]{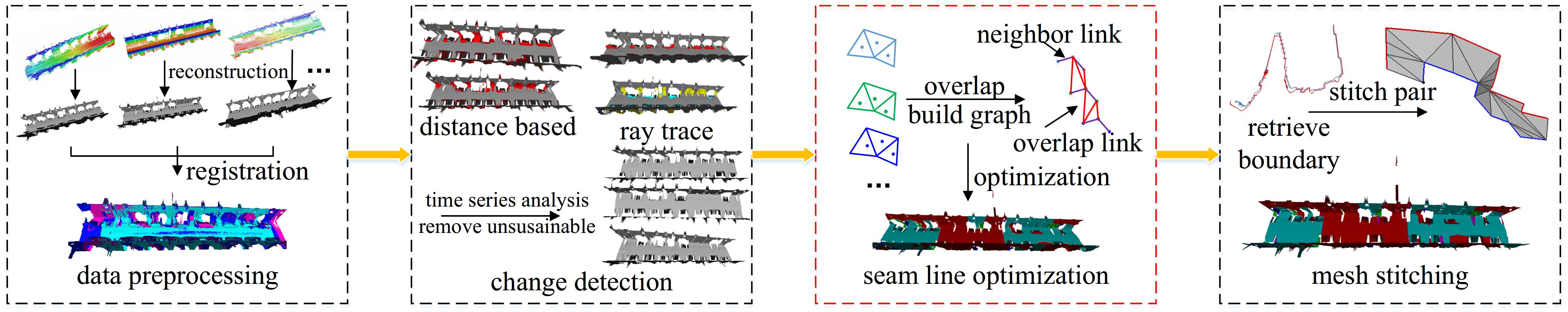}
	\caption{Pipeline of our mesh update method. \textbf{Step 1} preprocessing: mesh generation and registration; \textbf{Step 2} Change detection: combine distance based and ray tracing based method detect changed objects, time series based sustainability analysis to keep consistent objects; \textbf{Step 3} mosaicking: quadratic pseudo boolean optimization based seam line optimization; \textbf{Step 4} stitching: stitch the seam lines.}
	\label{Fig.workflow}
\end{figure*}

%There are temporary or permanent changes in long term mapping. There are two levels of temporary changes, moving during the collecting, such as pedestrians, and moving between the two collection period, for example, parking cars. For permanent changes, they will be updated in the map.
%For the input dataset, in the first situation to remove moving objects, a single session is required, and multi sessions are needed for the second situation.
%For the first situation, image or point cloud sequence analysis are the main method.
%Motion tracking in 2D images and 3D space are adopted \cite{jiang2017incomplete}. For single session, low dynamic objects are difficult to be identified.
%For multi-session, some low dynamic objects can be detected.
%Most works treat multi-session data as one large session, through the inconsistent analysis same with single session, and all the processes are based on point cloud in the procedure. 
%In this situation, it is not a update pipeline, and if we want to add a new collection, the whole workflow, including location and change detection, needs to be run again, and this takes a lot of time.
This paper addresses the issue of change detection, sustainability assessment and map updating from heterogeneous (geometric) data with arbitrary quality (resolution and precision) coverage/viewpoint and acquisition dates. The updated mesh is a mosaic of the input meshes optimally exploiting their heterogeneity. The pipeline is shown in \Cref{Fig.workflow}.
%In the paper, we produce an automatic pipeline to update mesh map in 3D, we concentrate on the sustainable changes using several collections from different times, and we handle this in an update way, after meshed generation, only the mesh and sensor position is needed. 
%Our workflow focus on sustainable update, including change detection analysis and map update.

%There are three steps in the framework:
%\textbf{Time series based change detection on meshes:} sustainable objects 
%\textbf{Seam line optimization:}
%\textbf{Seam line stitch:}

\section{RELATED WORK}

Long term localization \cite{sattler2018benchmarking} and mapping \cite{dymczyk2018long} are research topics in both robotics and computer vision communities. For long term mapping, the work presented here is related to three research areas as listed below:

\paragraph{3D change detection} is widely researched \cite{qin20163d}. For mobile mapping, there are two levels about change analysis, motion analysis along the sequences and changes detection during a long time. For motion analysis, the input data is a sequence, and for change detection, the inputs are several sequences from different times. For point cloud sequence, 3D Vector field based object analysis is used to get the motion and static objects \cite{jiang2017static}.
Ray tracing is used to define dynamic and motion analysis based on iterated closest point is used to obtain the velocity \cite{pomerleau2014long}.
Ray tracing and Dempster–Shafer Theory are used for change detection in \cite{xiao2015street}. In indoor environment, low cost RGB-D sensors are used for long term mapping, extend truncated signed distance function (TSDF) to handle changes during reconstruction\cite{fehr2017tsdf}.
Focus on point cloud analysis, occlusion and change analysis is proposed for long term mapping, and clustered objects are used to analyze, instead of single point\cite{ambrucs2014meta}.

\paragraph{Map update} For 2D map update, considering seam line, difference geometry and illumination between different image, seam line optimization is widely researched \cite{lin2015adaptive}. But in 3D space, the problem becomes complicate, points or triangles are not grid and overlap is not regular. Point cloud is used to describe the map in most work \cite{miksik2019live}. And it is easy to update the map, add the new point cloud to the old point map. After all the dynamic objects are removed, the mesh can be generated \cite{fehr2017tsdf}. In SLAM application, moving object is also an issue, use deep learning method to detect and remove dynamic objects in LiDAR-based SLAM \cite{chen2019suma}, semantic mapping based on Recurrent-OctoMap is proposed for a long-term semantic mapping on 3D-Lidar Data\cite{sun2018recurrent}, and boosted particle filter based change detection method is applied on definition digital maps \cite{pannen2019hd}. 

\paragraph{Mesh stitching} is a recurrent need in 3D model reconstruction, due the necessity to combine several individual data acquisitions to cover a whole scene. The Zippering algorithm \cite{turk1994zippered} can be used to obtain a seamless representation with no overlap.
To handle large scale reconstruction, after 3D constrained Delaunay triangulation, Graphcut is used to fuse several meshes \cite{vu2011large}, but the result should be watertight, at the same time, the overlap area should be detected and the buffer is the non-overlap area.
Joint alignment and stitching of non overlapping meshes method is proposed to stitch two part of meshes which do not have overlap areas \cite{brandao2012joint, brandao2014three}, after aligned, the vertices are linked using order assignments.
Hole filling is also a way to link vertices to obtain a whole mesh \cite{liepa2003filling}.

The previous work most related to ours \cite{wu2020moving} tackles the three issues and proposes a pipeline similar to ours, but the focus is on scene reconstruction and the objective is simply to aggregate LiDAR point collected by two 2D range scanners with the same quality mounted on the same platform and acquired during a single acquisition. Our contributions relative to this work and the more specific works cited above can be summarized as follows:
\begin{enumerate}
	\item A mesh based 3D change detection method based on combining distance and ray tracing criteria.
	\item A time series analysis to define and assess the sustainability of changes on 3D meshes.
	\item A global quadratic pseudo boolean optimization(QPBO) framework to create a mesh mosaic maximizing novelty and quality and minimizing seam lines that are then stitched by triangle strips. 
\end{enumerate}

% semantic classification is also a change detection method
% sustainable define
% the problem is that sustainable is not widely used
% think the difference between static and sustainable
% \subsection{Sustainable update}

\section{CHANGE DETECTION}

\subsection{Preprocessing}

The input of change detection method should be well registered surface meshes with viewpoint information: either a single viewpoint for all triangles (depth images, image space dense matching or fixed panoramic Lidar) or an optical center per point(Mobile, Aerial or Drone Lidar Scanning). In practice we used sensor meshing, creating triangles based on the pixel grid structure for image based point clouds or the regular scan structure (creating triangles based on successive points and successive scanlines) for Lidar (most Lidar sensors give access to this information which is can be preserved during the export). Because Lidar scans are at a very hugh frequency (typically ~300kHz), the three viewpoints for each vertex of a triangle are always very close so choosing their barycenter is a good approximation to provide a triangle viewpoint.
% time series part
%Moreover, our method requires a timestamp for each triangle, that can be defined for the whole mesh (for dense matching meshes for instance) or per triangle (for example, MLS meshes acquired continuously in time).

%In geometry change detection, when using time series meshes, the non-sustainable objects can be analyzed using change detection method base on time analysis. Sustainable changes are the objects which should be updated in the map.
%Time series analysis method on meshes is proposed. Visibility based method and distance base method are combined in the algorithm. 
%Distance based method can only know the difference between two meshes, but it could not tell it is change area or not. In 3D space, the overlap may be different because of coverage and occlusion, the different place is changed or not, and a further decision is needed.
%Visibility based method is sophisticate, and it can classify change areas and consistent areas ideally. Ray intersection is widely used with the visibility information. Considering the mesh is continue in 3D space, a tetrahedron based intersection is adopted. In practice, we find that when the intersection angle is small(nearly parallel), there will be a gross error in intersection. To overcome the shortage of calculation error, distance based method is a good supplement.

\subsection{Consistency analysis}

\begin{figure}[t]
	\centering
	\includegraphics[width=0.8\linewidth]{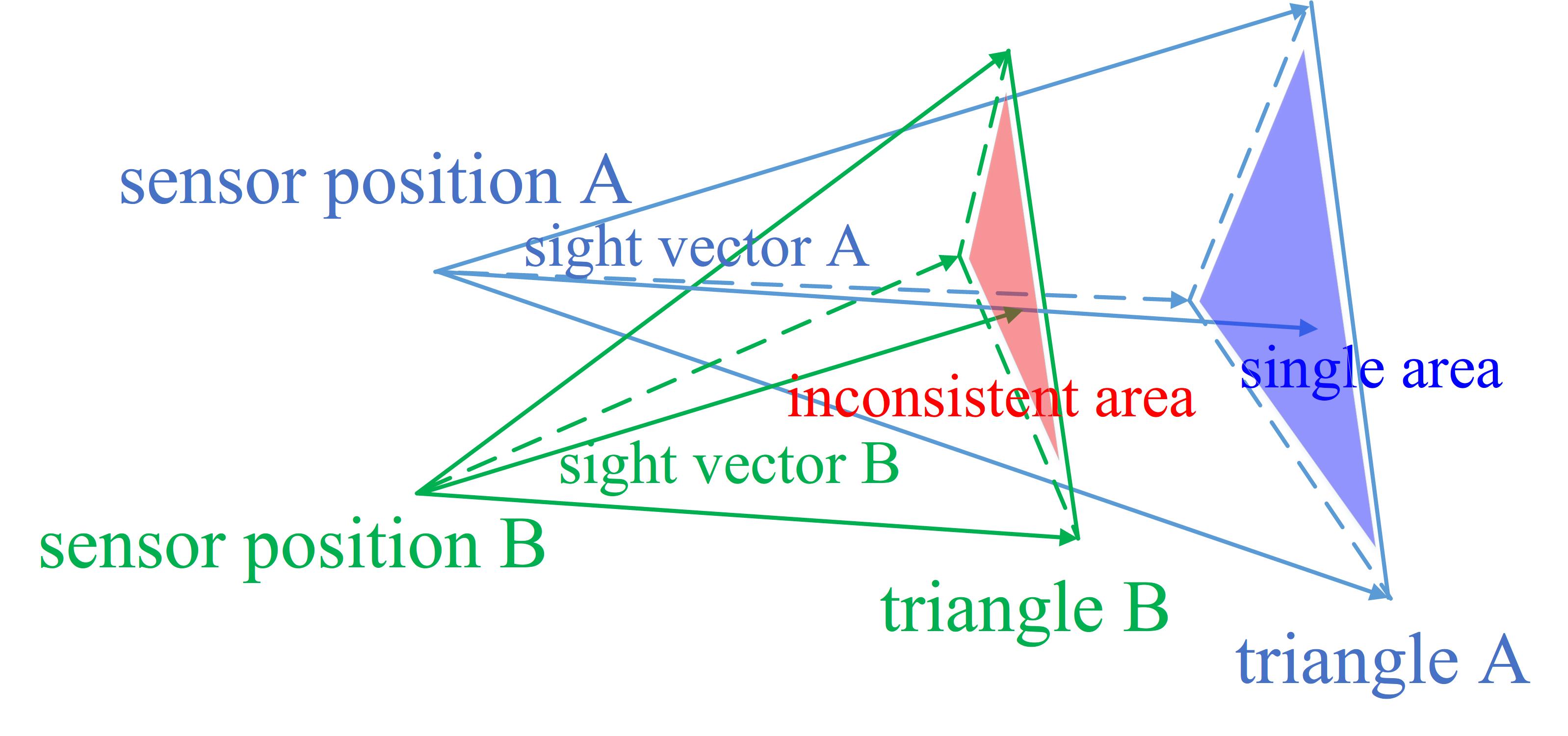}
	\caption{Asymmetry of 3D change detection: triangle B intersects empty tetrahedra A indicating a conflicting volumetric information. If B is older than A, it means B belongs to a surface that has disappeared. In the opposite case, there is no conflict as B simply belongs to a new surface that was not exist when A was acquired.}
	\label{Fig.unsymmetric}
\end{figure}

\begin{figure}[t]
	\centering
	\subfigure[Sensor information based visibility analysis.]{
		\label{Fig.Change_3d:b}
		\centering
		\includegraphics[width=0.8\linewidth]{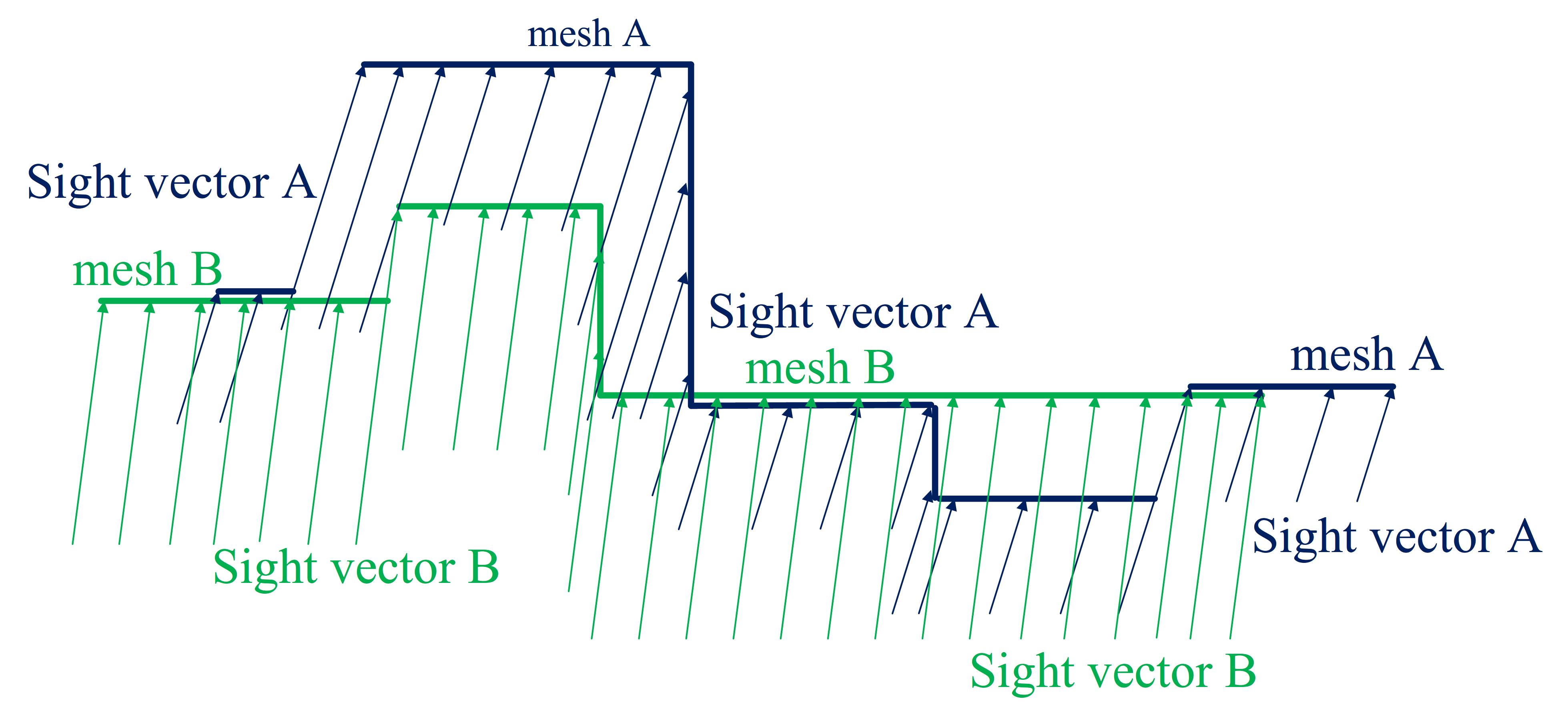}
	}
	
	\subfigure[All possible situations in change detection.]{
		\label{Fig.Change_3d:c}
		\centering
		\includegraphics[width=0.9\linewidth]{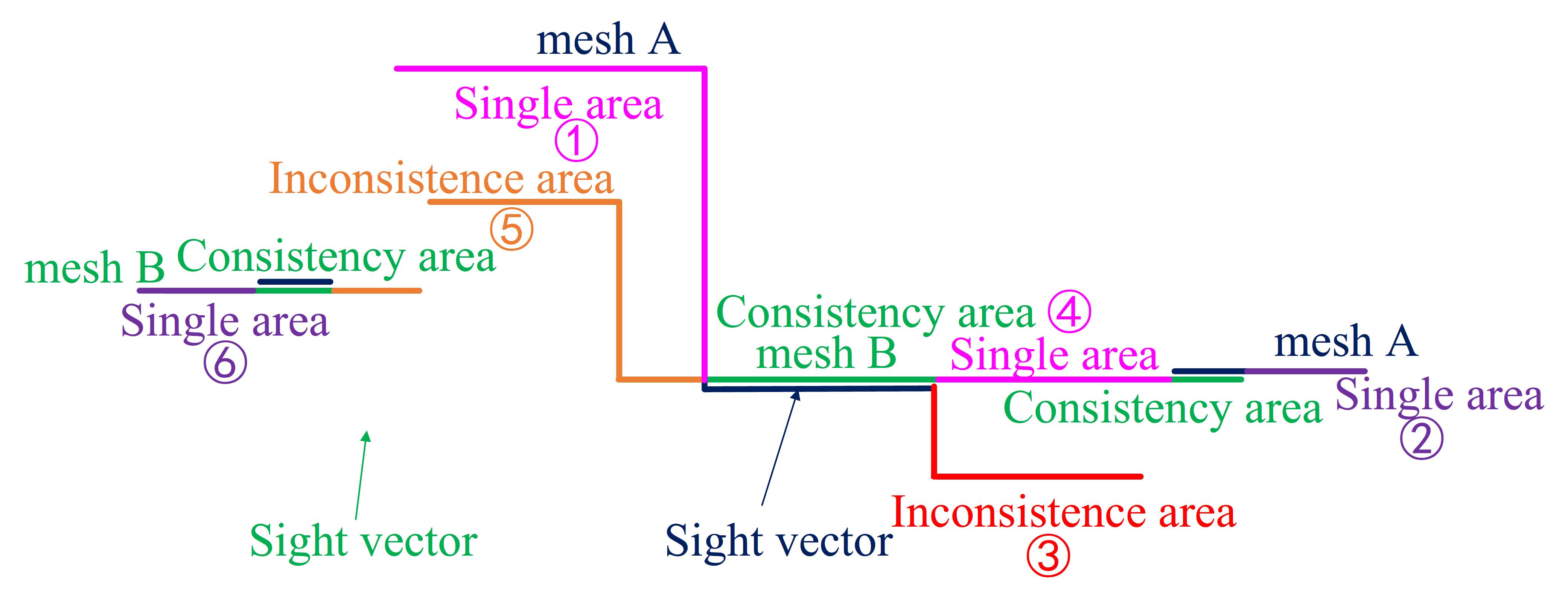}
	}
% quality is not concerned.	
	\caption{A 2D profile view(from above) of change detection analysis.} % and map merging for an old mesh with high quality and a new mesh with lower quality.}
	\label{Fig.Change_3d}
\end{figure}

Change detection in 3D space is more complex than in 2.5D space: in 2.5D (depth map on a regular grid), if depths disagree between the olde and newer depth maps, they are conflicting and the newest is kept. In 3D, using viewpoint information, each triangle of each mesh defines an empty volume of space (the tetrahedras formed by the triangle and its viewpoint), and intersection of a tetrahedra of one mesh with a triangle of the other can be conflicting or not, as shown in \cref{Fig.unsymmetric}, such that conflict is an asymmetric relation. Note that because our inputs may have different resolutions, intersecting triangles with rays is not a robust conflict assessment as a small triangle could simply fit between rays, which justifies out choice for tetrahedra based visibility assessment. This justifies our choice for to define consistency for each triangle of each mesh by three possible cases illustrated in \cref{Fig.Change_3d}:
\begin{enumerate}
    \item consistent: the triangle is close enough to a triangle of the other mesh. This is computed by thresholding (Kd tree accelerated) distance computation, and the list of close enough overlapping triangles from the other meshes is stored for each triangle. 
    \item conflicting: the triangle is not consistent and intersects the empty space of at least one newer triangle from another mesh, as areas \circled{3} and \circled{5}.This is computed by ray tracing on non consistent triangles.
    \item single: the triangle is neither consistent nor conflicting, meaning it has no counterpart in any other meshes, as areas \circled{1}, \circled{2}, \circled{4} and \circled{6}. This is also computed by ray tracing
\end{enumerate}

%This assumes that the information of the sensor position is not lost during the acquisition, so for each triangle, we know where it was acquired from so we can build the sight ray from sensor to triangle vertices (a very thin tetrahedron based on the sensor position and the 3 triangle vertices). As shown in \cref{Fig.Change_3d}, this allows to define the inconsistent situation as that a triangle from dataset A intersecting a ray from dataset B, as this indicates that space at the intersection was occupied in A and empty in B. Same like 2D map update, \emph{consistent} area, \emph{inconsistent} area, and \emph{single} area are used to define the different situations.

For conflicting situation, if mesh A is collected after mesh B, only delete triangles which correspond to the orange area \circled{5}, and area \circled{3} is kept in \Cref{Fig.Change_3d:c}.

%However, contrary to the 2D case, this situation is not symmetrical anymore, as in that case the ray from A does not intersect the triangle from B, area \circled{1} in \cref{Fig.Change_3d:c}. Consequently, we only delete the triangle from A if it is collected after the triangle from B (but not the opposite), which corresponds to the orange area \circled{5}. On the other hand, the red area \circled{3} is different with area \circled{5}, and area \circled{3} is a new object to be updated. In the two inconsistency situations, there are areas labeled as \emph{single}, pink areas \circled{1} and \circled{4}. 
%In the coverage aspect, there are single areas, i.e. purple areas \circled{2} and \circled{6}. For the single area, it is same with the 2D case, all the data is kept.
%For the consistent case, things are symmetrical and work the same as the 2D case: if triangles from A an d B are close enough (based on a distance threshold $D_t$), then they are redundant so we remove the triangle with lowest quality, which corresponds to the dark blue and green areas in \cref{Fig.Change_3d:c}.

% sustainable property 
% here is important, the contribution contains this step
\subsection{Time series analysis}
\label{sec:timeseries}
%we define a property for each triangle, records the first collected time and last collected time. During the ray tracing, when there is a conflict, we can compare the last observation time to know the triangle is newer or older, so the method does not depend on the input order.
To analyze time series, we update time information of each triangle incrementally by processing our input meshes in acquisition order. Triangles conflicting with a newer acquisition are systematically removed as they correspond to an object that has disappeared. In the other cases, we simply list for each triangle all the consistent triangles from the other meshes, and the time information of consistent triangle is updated.
We propose to assess the sustainability of a new object by requiring it to be confirmed by a new data source more than the sustainability threshold $T_s$ after its first appearance in the data. After all meshes are processed, for each triangle, we know the time of first appeared, and the time of last observation. Thus we keep the \textit{sustainable}  triangles which are consistent for time $T_s$.
% not readable
%Thus we tas as \textit{sustainable} the triangles for which at least one consistent triangle is newer by at least $T_s$.

% illustrated by WT
%During pair based consistency analysis, the conflict areas are removed and a time span of consistent triangles is generated. 
%Then the third mesh is compared to both the old meshes. After all meshes is processed, each triangle has a time span, and the triangle is remained only the time span is longer than a certain time(a week).

%We consider that an change is sustainable if it remains stable for a long time.
%After each change detection step, changed areas are removed. Then the change detection methods are used iteratively.
%If the data is collected with time sequence, the dynamic objects can be detected.
%In our situation, after change detection, we only update the consistent objects which can last a long time. In the experiment, the shortest period is one week.

\section{3D MESH MOSAIC}

In this section, we aim at creating a mesh mosaic from the sustainable triangles defined in the previous section(cf \Cref{sec:timeseries}).
%\st{as defined above so all triangles mentioned are sustainable.}
%mosaic on 3D mesh also emphasizes on searching seam lines. %A mosaic method on several meshes is proposed, and it contains finding and stitching seam lines.
% it is important to illustrate the selecting triangles

\subsection{Seam line optimization}

2D mosaicking is a widely used method to stitch several images (possibly resampled) in the same geometry into a larger one in a way that minimizes radiometric jumps across seam lines \cite{lin2015adaptive}. To extend this concept into 3D meshes, the main issue is that 3D triangles do not coincide exactly as 2D pixels do, so what we want to keep as many non overlapping triangles as possible while minimizing the length of seams in the final mosaic. So while 2D mosaicking is a pixel labeling problem, 3D mosaicking is a boolean optimization problem to decide whether we keep each triangle or not.
In our method, we construct a graph where nodes are triangle and having two types of edges: consistency edge between consistent triangles (of different meshes) and neighboring edges between neighbor triangles (of the same mesh). In order to obtain a mesh with maximum triangle area without overlap, and minimum seam lines, we minimize: 
\begin{equation}
E(\mathbf{x}) = \sum_{p \in \mathcal{G}}\phi_p(x_p) + \lambda \sum_{(p, q) \in \mathcal{N}}\psi_{pq}(x_p, x_q)
\label{equation:standard_opt}
\end{equation}
In \Cref{equation:standard_opt}, $\mathbf{x}$ is the label result, $\mathcal{G}$ is the graph, $\mathcal{N}$ is neighbor, $q$ and $q$ are the nodes in the graph. For the data term, in order to keep maximum triangles and less minimum seam line, the nodes are divided into two types: boundary triangles and normal triangles. We keep less boundary triangles and more normal triangles.
For the smooth term, there are two type of link between the nodes: triangle from the same mesh and triangle from different meshes.  For the first situation, the neighbor triangles should have the same label; for the second situation, to avoid the overlap, these neighbor triangles should not selected both.
To expand the formula with information from meshes, we write the mesh mosaic problem as minimizing optimization:
\begin{equation}
\label{equation:obj_simplify_general}
\begin{split}
E(\mathbf{x}) = & \lambda_1 \sum_{M_i} \sum_{T^i_j \in M_i} Q(T^i_j) x_{i,j} + \lambda_2 \sum_{M_i} \sum_{T^i_j\in \mathcal{B}_i} L(T^i_j) x_{i,j}\\
+ & \lambda_3 \sum_{(T^{i_1}_{j_1},T^{i_2}_{j_2})\in \mathcal{C}} x^{i_1}_{j_1} \cdot x^{i_2}_{j_2} \\
+ & \lambda_4 \sum_{M_i} \sum_{(T^i_{j_1},T^i_{j_2})\in \mathcal{A}_i} L(T^i_{j_1},T^i_{j_2}) \cdot XOR(x_{i,j_1},x_{i,j_2})
\end{split}
\end{equation}
To define the optimization problem proposed, in \Cref{equation:obj_simplify_general}, we will use the following notations:
\begin{itemize}
    \item $x_i^j \in \{0,1\}$ is a boolean label on each triangle $T^i_j$ of mesh $M_i$, indicating if the triangle is \textbf{kept} (1) or \textbf{removed} (0) in the mosaic. %$\mathbf{x}$ is a vector concatenating all the $x_i^j$.
    \item $Q(T^i_j)$ is the quality of triangle $T^i_j$, this can be a constant number, triangle area size, etc.
    %The triangle area is used in the experiment.
    \item $\mathcal{B}_i$ is the set of triangles of mesh $M_i$ with at least one boundary edge. For a triangle $T_i^j\in \mathcal{B}_i$, we call $L(T_i^j)$ the length of its boundary edge(s).
    \item For two triangles $T^i_{j_1},T^i_{j_2}$ of the same mesh $M_i$ sharing an edge, $L(T^i_{j_1},T^i_{j_2})$ is the length of their common edge.
    \item $\mathcal{C}$ is the set of all consistent pairs of triangles from two different meshes, which means they are below the distance threshold and that they overlap.
    \item $XOR(x_1,x_2)=x_1+x_2-2x_1 \cdot x_2$ the exclusive \textit{OR} logical operator.
\end{itemize}

where $\lambda_1, \lambda_2, \lambda_3$ and $\lambda_4$ are the weight balance, $\lambda_3$ is a large constant discouraging overlaps.
%The first two terms ensure that the result is a mosaic, in the sense that it takes as many input triangles as possible provided that they do not overlap, and it encourages to keep triangles of highest quality on overlaps.
%The last two terms count boundary length and is split between boundaries of the input meshes that are still part of the output mosaic and new boundaries caused by deleting triangles.
The first two terms are the extension of data term, and the lats two terms are the extension of smooth term in \Cref{equation:standard_opt}. The first term ensures that it takes as many input triangles as possible, and it encourages to keep triangles of highest quality. The second term counts the boundary triangle from the mesh, and we want to keep fewer boundary triangle, the third term make triangles from different mesh should have different labels, i.e. no overlap, and the last term encourage the result has less seam lines.
A straightforward approach to the definition of quality is through resolution: higher resolution means more information on a given area in a scene.
In the framework, seam lines and overlaps should be minimized, quality should be maximized ($\lambda_1$ is minus sign). Choosing a constant quality measure will naturally favor higher resolution as it will favor maximizing the number of triangles, thus the higher resolution mesh can be obtained.
However other choices can be made for the quality metric to take into account other factors such as precision or certainty.
Note that in our framework, quality has the priority over novelty only on parts of the scene that have not changed, but novelty has priority by construction on changing parts of the scene as the optimization runs on sustainable triangles only.

\subsection{Seam line stitching}

After seam line optimization, the result is a set of individual mesh parts (connected components of triangles selected by the seam line optimization). Thanks to the optimization minimizing seam length, these parts are quite compact, but they are not continuous across seams. To obtain a continuous mesh, stitching the parts across seams is necessary, which is performed in two steps: match the boundaries of each connected component, and then link the adjacent boundaries, more detail can be found in our previous paper \cite{wu2020moving}.

% to do: an experiment consider both dynamic objects inner session and between sessions
% fusion, mosaic, stitch
\section{EXPERIMENTS}

We conducted our experiment on Robotcar dataset \cite{maddern20171,maddern2020real}, the laser sensors are mounted on the mobile car, and a long time series dataset is collected. Meshes can be generated from LiDAR point cloud from a 2D sensor, based on sensor topology \cite{boussaha2018large}, and the sensor position can be retrieved after reconstruction.
After mesh generation, meshes are registrated using global ICP method \cite{yang2015go}.

\subsection{Change detection}
For Robotcar dataset, because the 2D range sensor is mounted on the bottom of the car, the ray to the road edge is nearly parallel to the road plane which bring large errors to the intersection in ray tracing.
To overcome the shortage of intersection, the distance base method and visibility based method are combined. 
The distance based method is symmetrical, but the visibility analysis is not symmetrical. To make the method symmetrical, each triangle will store a time stamp, triangle is kept or removed depends the time and change detection result, so change detection method doesn't rely on the input order, both meshes are processed symmetrically. In the experiment, the occlusion is decided using the timestamp of the two triangles.

\begin{figure}[t]
	\centering
	\subfigure[Result of distance base method on mesh \textbf{$t_0$}.]{
		\label{Figure.robotcar_change_resut:a}
		\centering
		\includegraphics[width=0.45\linewidth]{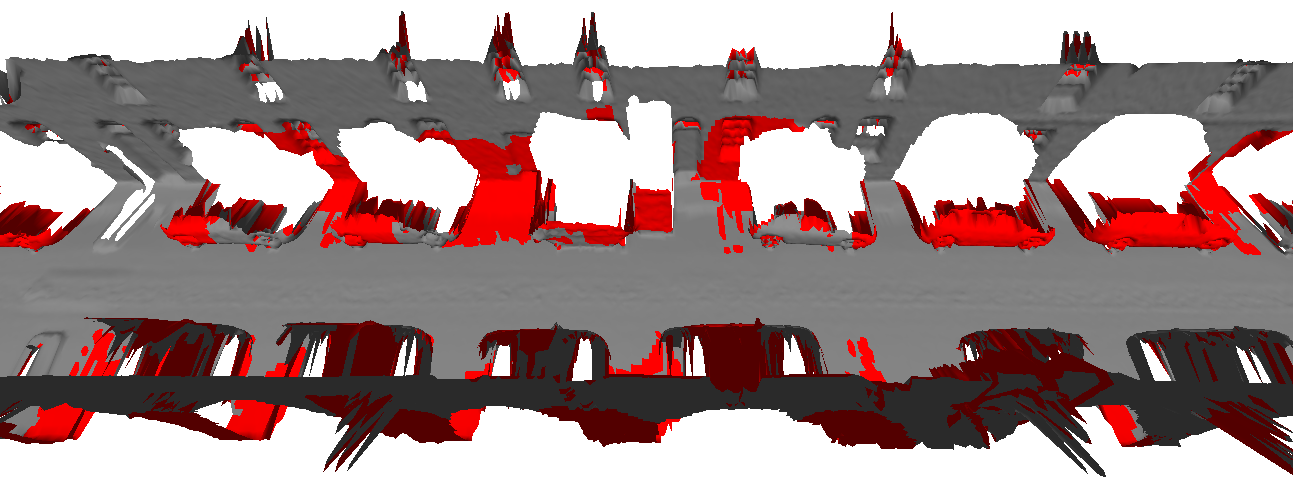}
	}
	\subfigure[Result of distance base method on mesh \textbf{$t_1$}.]{
		\label{Figure.robotcar_change_resut:b}
		\centering
		\includegraphics[width=0.45\linewidth]{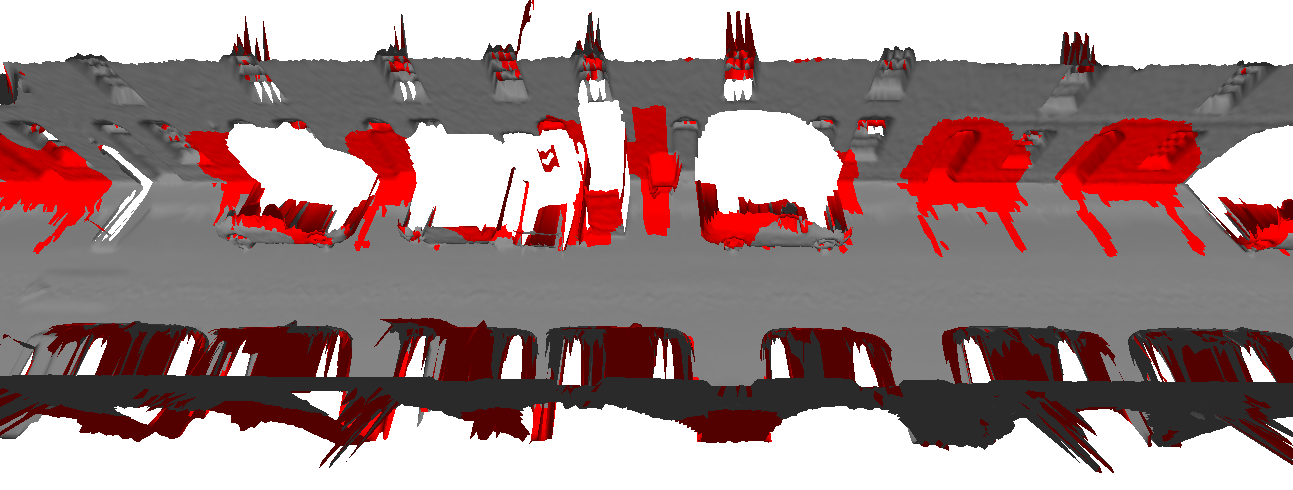}
	}
	
	\subfigure[Result of ray trace base method on mesh \textbf{$t_0$}.]{
		\label{Figure.robotcar_change_resut:c}
		\centering
		\includegraphics[width=0.45\linewidth]{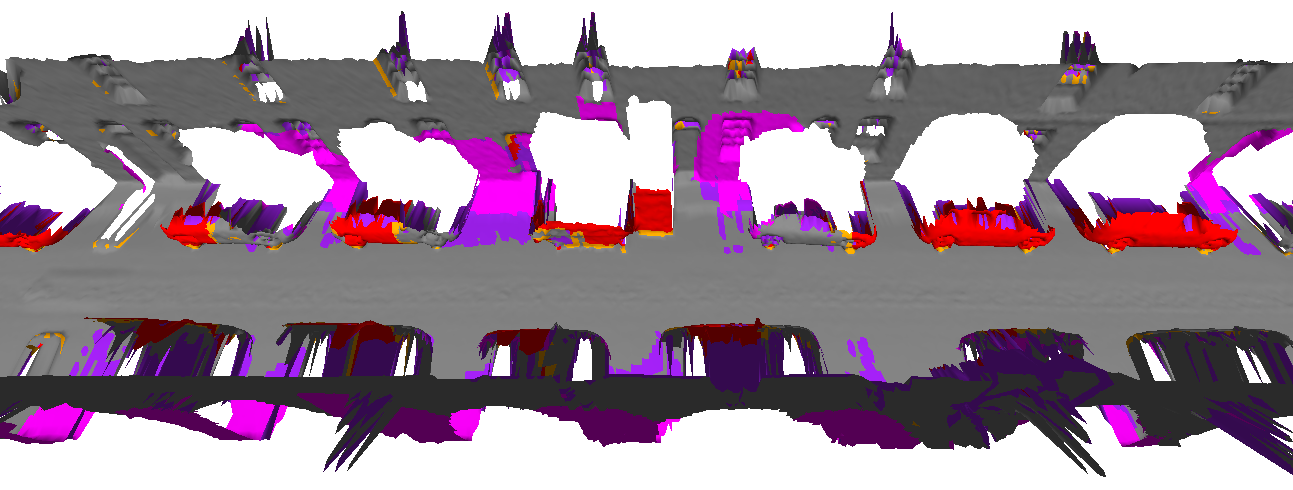}
	}
	\subfigure[Result of ray trace base method on mesh \textbf{$t_1$}.]{
		\label{Figure.robotcar_change_resut:d}
		\centering
		\includegraphics[width=0.45\linewidth]{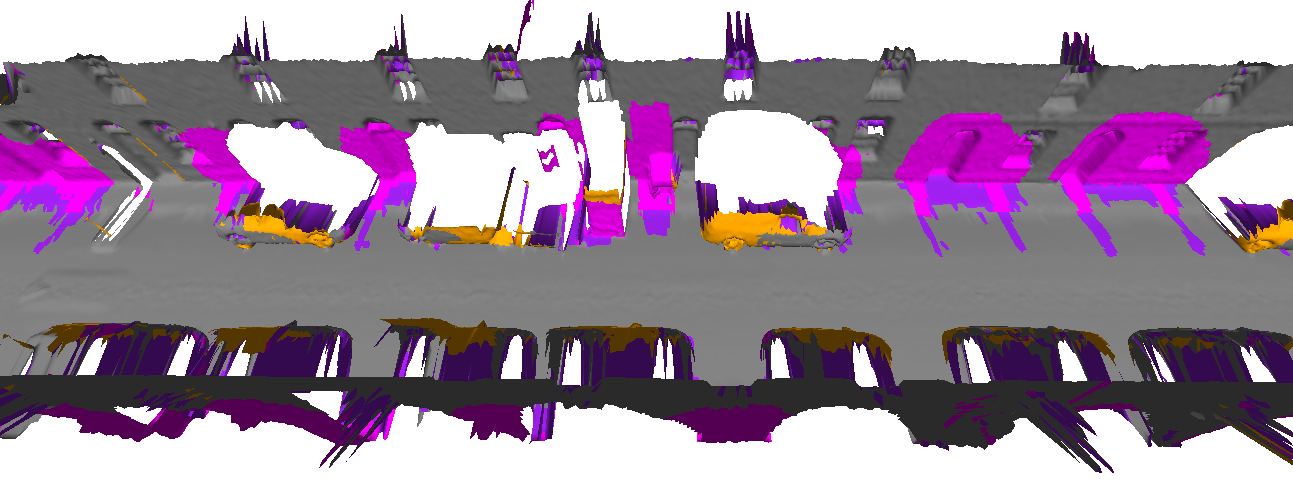}
	}
	\caption{Change detection result for Robotcar dataset. In (a) and (b), different area in\legendsquare{red}. In (c) and (d), same colors are used with \cref{Fig.Change_3d:c}, except consistent area in\legendsquare{gray}. Single area because of occlusion in\legendsquare{magenta}, because of overlap in\legendsquare{RoyalPurple}. Inconsistent area from old mesh is in\legendsquare{red}, from new mesh is in\legendsquare{orange}.}
	\label{Figure.robotcar_change_resut}
\end{figure}

In \cref{Figure.robotcar_change_resut}, using distance based method can only know the differences, as shown in \cref{Figure.robotcar_change_resut:a} and \ref{Figure.robotcar_change_resut:b}. Visibility analysis can tell the changed areas, as the red areas shown in \cref{Figure.robotcar_change_resut:c}. 
Because \textbf{$t_1$} is after \textbf{$t_0$}, single area because of occlusion in magenta in time \textbf{$t_1$} as shown in \cref{Figure.robotcar_change_resut:d}, we can not decide the areas are sustainable or not only with two inputs. After pair based change detection, time series analysis is used to know the consistent parts. 
%For the inconsistent part in \textbf{$t_1$}, as shown in \cref{Figure.robotcar_change_resut:b}, the inconsistent parts should be decided using time series analysis. 

% how to show the time series analysis?
% show the number of consistent triangles, remain triangles
%After pair based change detection, time series analysis is used to know the consistent parts. According to \cref{Fig.Change_3d:c},  after removing inconsistent areas, there are single parts. 
%The exist time is used to decide the sustainability, as shown in \cref{Fig.time_series_analysis}. Actually, the same area, the triangle number is influenced by the speed of the moving car. The numbers of triangles are convergent after several iterations.
% for the consistent triangle number, it is symmetric
%\begin{figure}[t]
%	\centering
%	\includegraphics[width=\linewidth]{figures/experiment/robotcar/showtriangle.png}
%	\caption{The triangle numbers along the iterated number.}
%	\label{Fig.time_series_analysis}
%\end{figure}

% also can compare with update model(keep the first observation)
\subsection{Seam line optimization}
\begin{figure*}[t]
	\centering
	\includegraphics[width=0.75\linewidth]{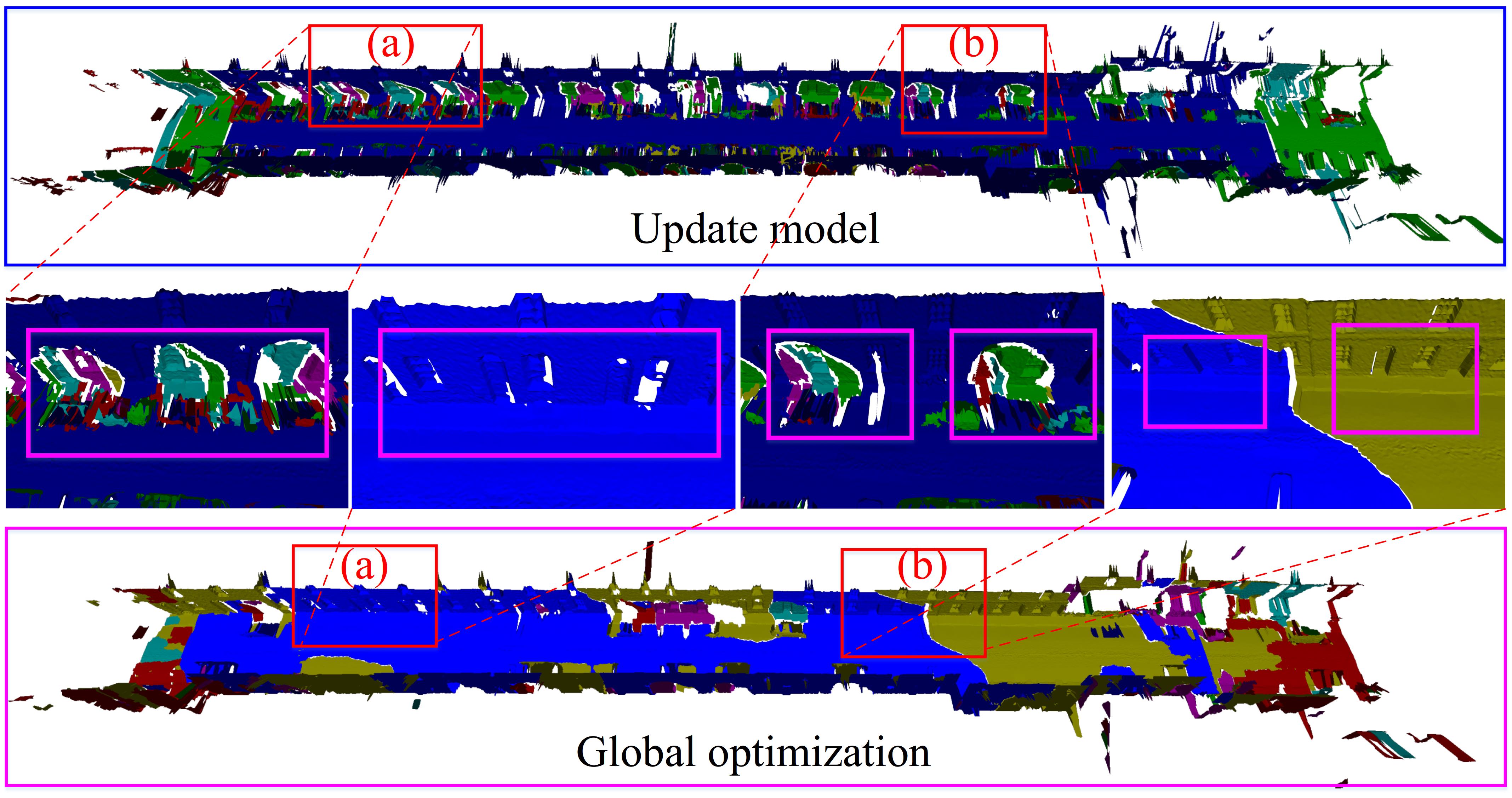}
	\caption{The triangle selection result of Robotcar dataset. In area (a) and (b), for update model, it only keep the first observation, the result depends on input order, usually result is filled by several parts. Our method can obtain a complete result.}
	\label{Fig.robotcar_opt_resut}
\end{figure*}

After change detection and time series analysis, unsustainable objects are removed, the seam line is optimized using QPBO \cite{rother2007optimizing}. A result is shown in \cref{Fig.robotcar_opt_resut}, the meshes which are from 7 meshes. 
To proof the effective of optimization method, an updated model is used for comparison. The updated model is common used in point cloud change detection \cite{ambrucs2014meta}. The updated model is just keeping the first observation.
As shown in \cref{Fig.robotcar_opt_resut}, the result of updated model depends on the input order of the mesh, the occluded hole is filled by several parts from different times shown in \cref{Fig.robotcar_opt_resut}. The optimization method selects a whole piece avoid increasing seam line shown in \cref{Fig.robotcar_opt_resut}.

% how to illustrate the importance of the seam line optimization?
% to show small parts?

% to do(discuss the solver?)
% result of the triangles total number of triangle, selecting triangles
% the different solvers
% it is confuse
% illustrate QPBO or QPBO + QPBO-I
%For the optimization method, there are two ways to solve the time series problem : update iteratively(divide and conquer) or solve the whole problem once. In the experiment, we found that if there are more than two meshes, some triangles are not labeled after QPBO, and QPBO with Improving(QPBO-I) is used to improve the label ratio. But QPBO-I is slow when there are a lot of triangles.

%A comparison of the results is shown in \cref{Table:solver_compared_large}, the iterated method obtains lower total energy, but the boundary energy and overlap energy is high.  In mesh mosaic, minimizing the boundary energy and overlap energy is more important, so the global solver is better than iterative method.
% use the time instead
%\begin{table}[t]
%	\caption{Comparison iteratively method and global solver method.}
%	\label{Table:solver_compared_large}
%	\begin{tabular}{|c|c|c|}
%		\hline
%		{\bfseries Factors} & {\bfseries Iterative method} & {\bfseries Global solver(QPBO-I)} \\
%		\hline
%		Run time($s$)  &  522.44 &  2107.27 \\
%		\hline
%		Energy  & -9599613 &  -9387984 \\
%		\hline
%		Triangles  & 295188 & 294534 \\
%		%\hline
%		%Area size($m^2$)  & 3932.67 & 3830.42\\
%		\hline
%		{\bfseries Boundary energy} & 1019123 & 976517 \\
%		\hline
%		{\bfseries Overlap energy}  & 24480 & 3120 \\
%		\hline
%	\end{tabular}
%\end{table}

% an experiment about resolution
\subsection{Mesh accuracy}
\begin{figure}[t]
	\centering
	\includegraphics[width=0.85\linewidth]{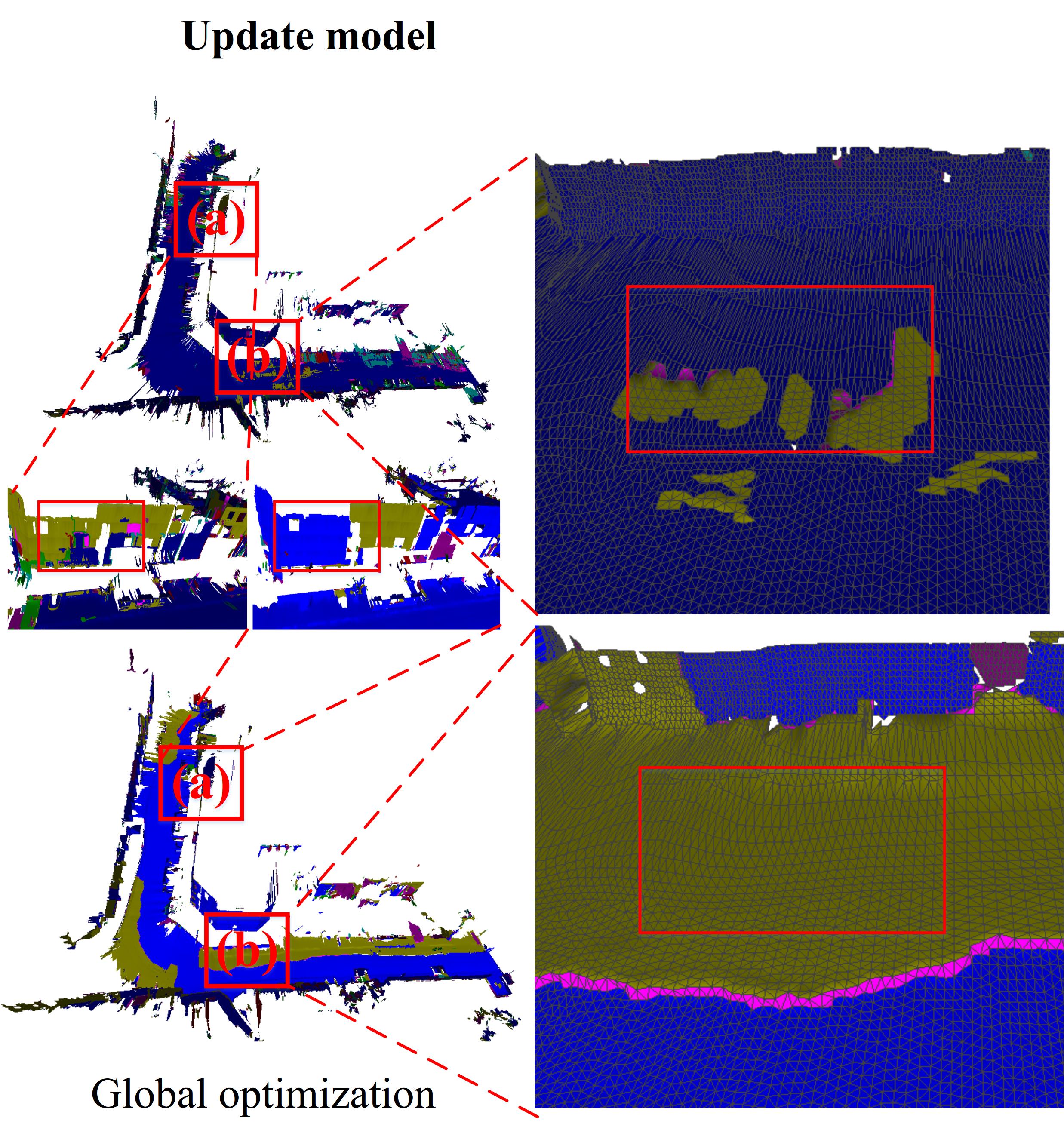}
	\caption{The detail of triangle selection result. In area (a), neighbor triangles are from same mesh, in area (b), our method obtains a result with higher accuracy. Different colors mean from different meshes. The triangle at stitch line is show in\legendsquare{magenta}.}
	\label{Fig.robotcar_accuracy}
\end{figure}

For Robotcar dataset, the used dataset is collected by a 2D sensor mounted on one car, when the car moves fast, triangles are large. Considering the property of 2D sensor(scan angle of the sensor has a big influence on point density), triangles on road boundary are large. Mesh accuracy(triangle size) is influence by the speed and trajectory of the car. For update model, result relies on the input order, and our method can select result with small triangles. As shown in \Cref{Fig.robotcar_accuracy}, area (a) means the out method obtain a result with less stitch area, and area (b) shows that even for the same area, the selection result can obtain a higher accuracy, triangles are smaller.

%other result
\subsection{Implement detail}
\begin{figure}[t]
	\centering
	\includegraphics[width=0.9\linewidth]{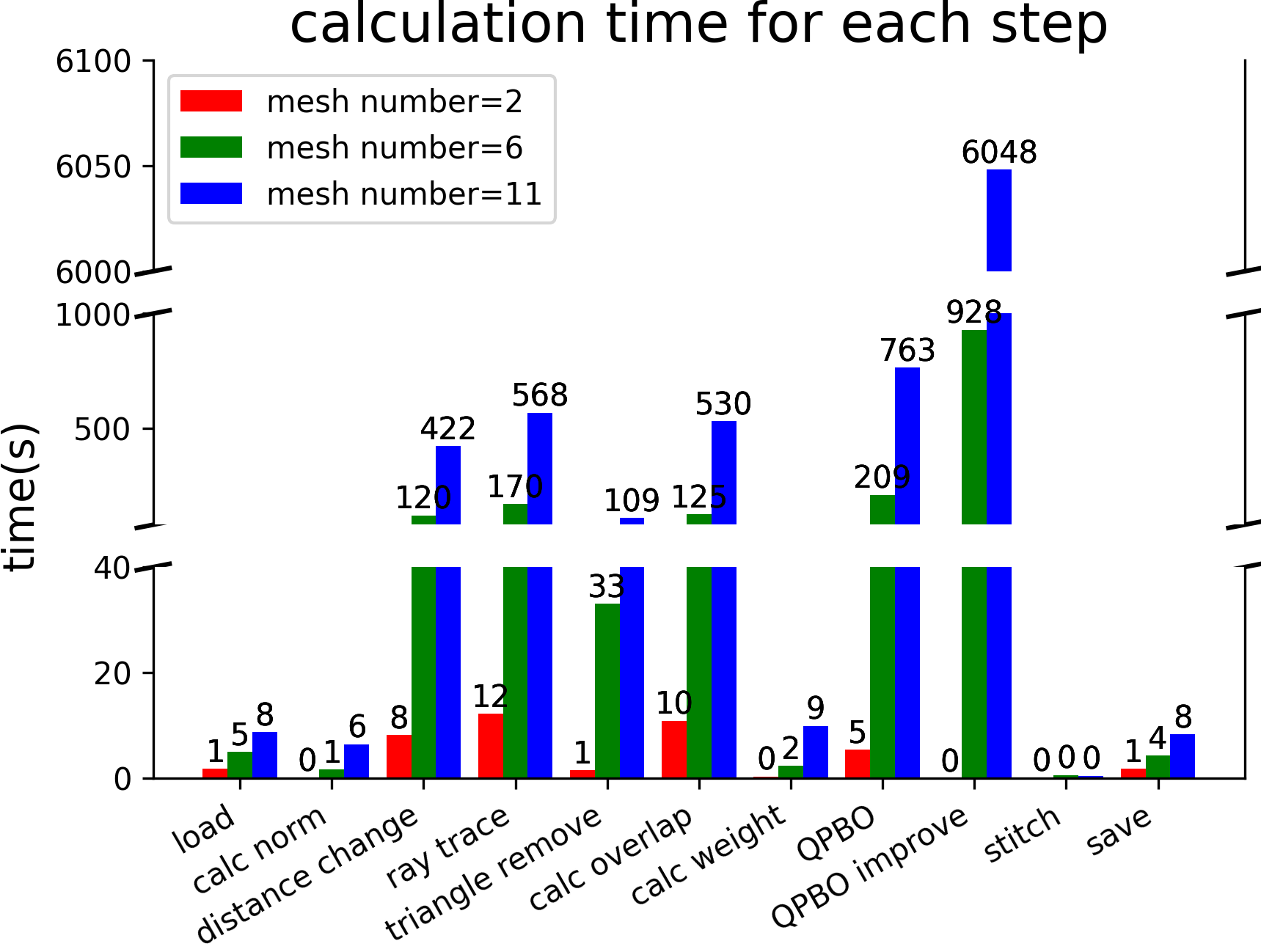}
	\caption{Computation time for each step with vary number of meshes. QPBO and QPBO improve optimization consume most of the time.}
	\label{Fig.runtime}
\end{figure}

The pipeline is implemented using c++, based on CGAL library \cite{cgal:eb-18b}. Computation speed is mainly influenced by the triangle number of the mesh and number of meshes. Considering there are several steps in the pipeline, the time for each step is shown. For the number is 2 or 6 means the first 2 or 6 mesh. The computer system is Ubuntu18.04, with 15.4G memory and CPU is Inter core i7-10510U. In \Cref{Fig.runtime}, if the number of triangles is large, then the QPBO improve optimization will take a long time, but when the number is 2, QPBO can solve the problem without improve method. Time $0s$ in \Cref{Fig.runtime}, means the running time is smaller than 1$s$. Seam line optimization step takes most to the time.

% a comparison with the original hole filling result
\subsection{Mesh stitching result}
\begin{figure}[t]
	\centering
	\includegraphics[width=0.87\linewidth]{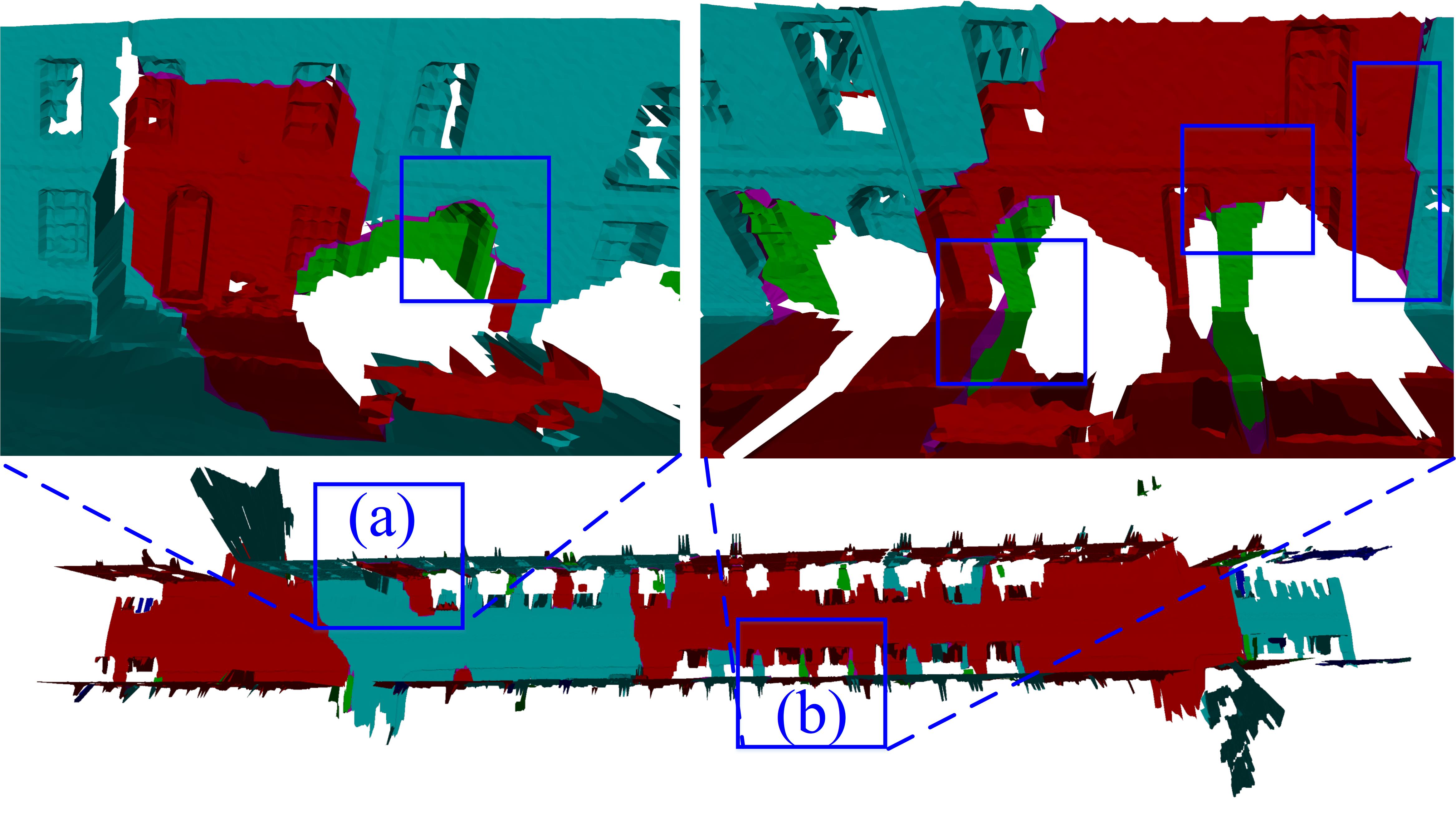}
	\caption{The stitch result of Robotcar dataset. Different colors mean from different meshes. The triangle at stitch line is show in\legendsquare{magenta}. Area (a) and (b) show the detail of stitching result.}
	\label{Fig.robotcar_stitch_resut}
\end{figure}

After seam line optimization, the mesh is made of by several parts. The boundaries are stitched to obtain a whole mesh, a result is shown in \cref{Fig.robotcar_stitch_resut}. Considering point distance, some vertices are merged to avoid small triangle.
% to do, a sub area?

\subsection{Stéréopolis dataset}
\begin{figure}[t]
	\centering
	\subfigure[Result of distance based method on mesh \textbf{$t_0$}.]{
		\label{Figure.stereopolis_change_resut:a}
		\centering
		\includegraphics[width=0.45\linewidth]{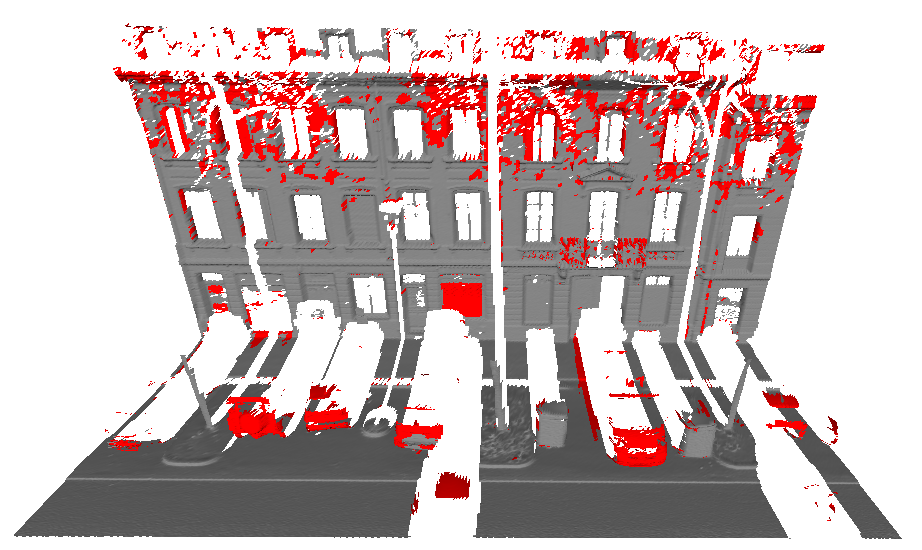}
	}
	\subfigure[Result of distance based method on mesh \textbf{$t_1$}.]{
		\label{Figure.stereopolis_change_resut:b}
		\centering
		\includegraphics[width=0.45\linewidth]{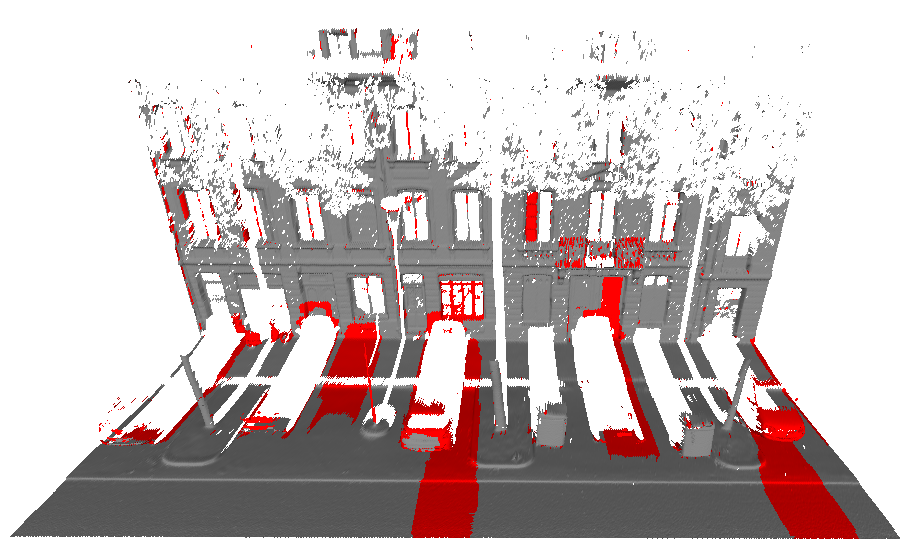}
	}
	
	\subfigure[Result of ray trace based method on mesh \textbf{$t_0$}.]{
		\label{Figure.stereopolis_change_resut:c}
		\centering
		\includegraphics[width=0.45\linewidth]{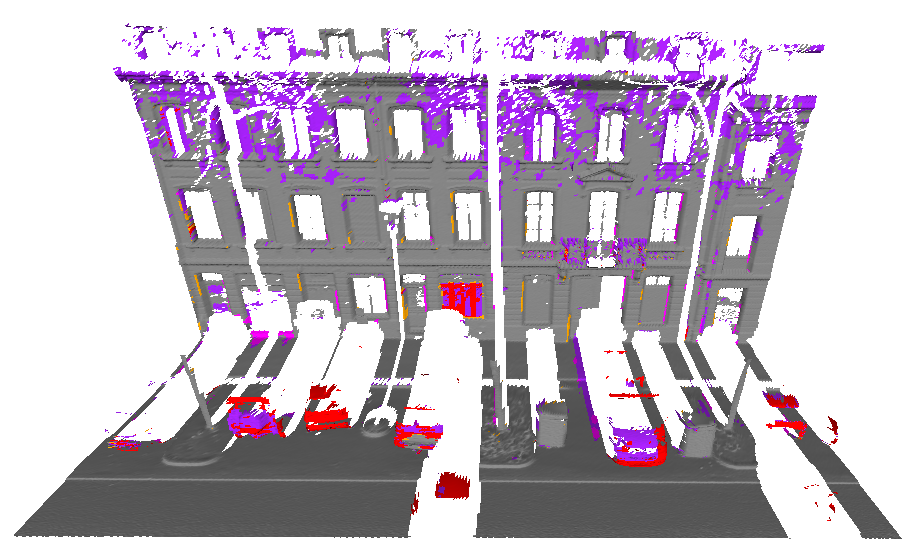}
	}
	\subfigure[Result of ray trace based method on mesh \textbf{$t_1$}.]{
		\label{Figure.stereopolis_change_resut:d}
		\centering
		\includegraphics[width=0.45\linewidth]{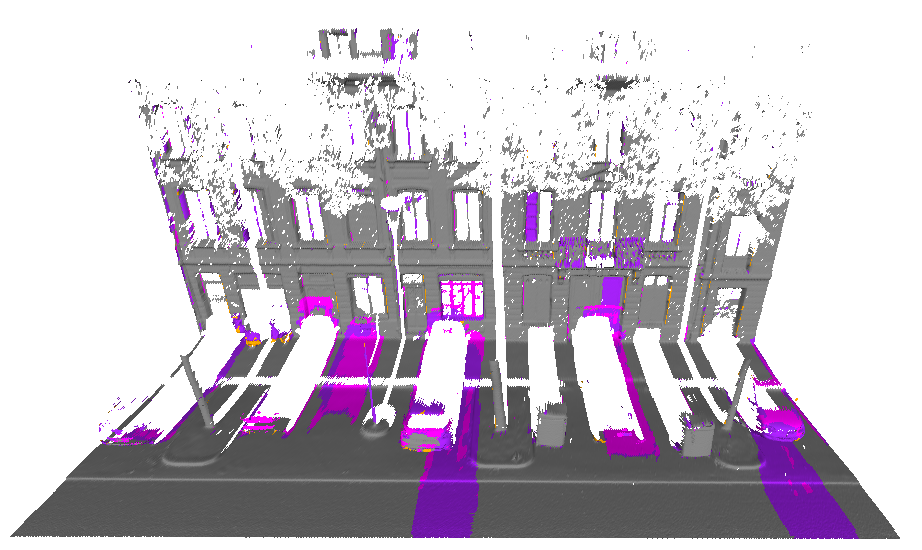}
	}
	\caption{Change detection result for stéréopolis dataset. In (a) and (b), different area in\legendsquare{red}. In (c) and (d), same colors are used with \cref{Fig.Change_3d:c}, except consistent area in\legendsquare{gray}. Single area because of occlusion in\legendsquare{magenta}, because of overlap in\legendsquare{RoyalPurple}. Inconsistent area from old mesh is in\legendsquare{red}, from new mesh is in\legendsquare{orange}.}
	\label{Figure.stereopolis_change_resut}
\end{figure}

Stéréopolis dataset \cite{paparoditis2012stereopolis} is also tested in the experiment. Due to the high accuracy of GPS and inertial measurement unit(IMU), the system produces a higher accuracy of the trajectory. 
Stéréopolis just collects same area dataset from two different times, in this situation, time series analysis can not be performed, to proof the third and forth steps of the workflow, we just removed the changed areas, the unsustainable objects are kept. 
As shown in \cref{Figure.stereopolis_change_resut}, the results of distance base method and ray trace base method are listed. Because there are only two sequences, the changed objects base on ray trace method are removed in the mesh \textbf{$t_0$}, and all the triangles in mesh \textbf{$t_1$} are kept. In the experiment, to avoid the influence of trees, in this example, the triangle in leaf areas are removed. 
Because the range sensor is on the top of the car, in the experiment, ray trace method obtains the same result as using both methods.
%4.1cm is the max width

\begin{figure}[t]
	\centering
	\includegraphics[width=0.80\linewidth]{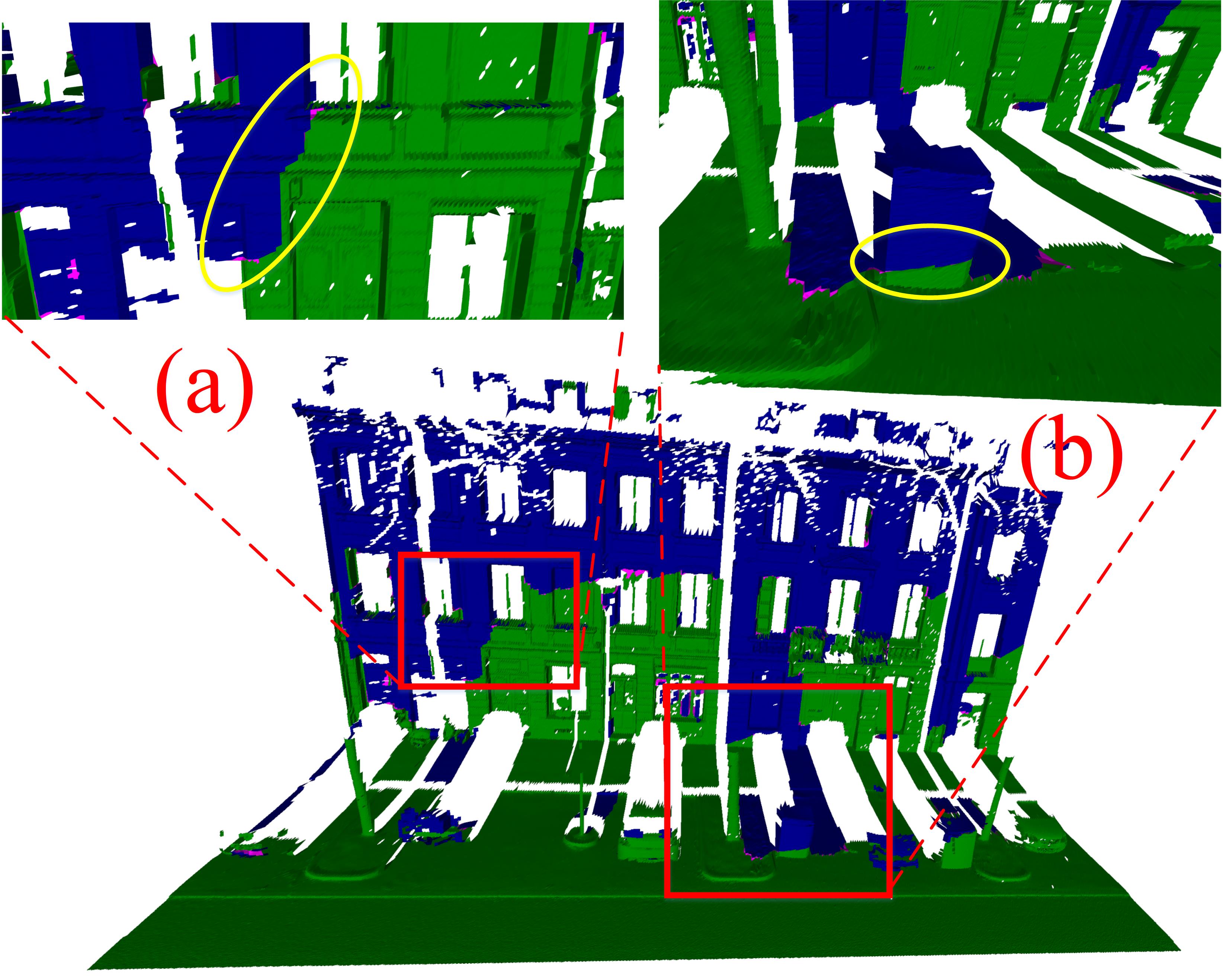}
	\caption{The stitch result of stéréopolis dataset, triangles from first mesh are shown in\legendsquare{BlueViolet}, triangles from the second one are shown in\legendsquare{OliveGreen}. The triangle at stitch line is show in\legendsquare{magenta}. Area (a) and (b) show that the boundary points are merged.}
	\label{Fig.stereopolis_stitch_resut}
\end{figure}

After seam line optimization, the two parts are stitched together, the result as shown in \cref{Fig.stereopolis_stitch_resut}. This example shows, the first mesh is more complete on the building facades and the second mesh is more complete on the ground. For the stitch step, the boundary line distance also considered, some points are merged as show in the yellow ellipses.

\begin{figure}[t]
 	\centering
 	\includegraphics[width=0.80\linewidth]{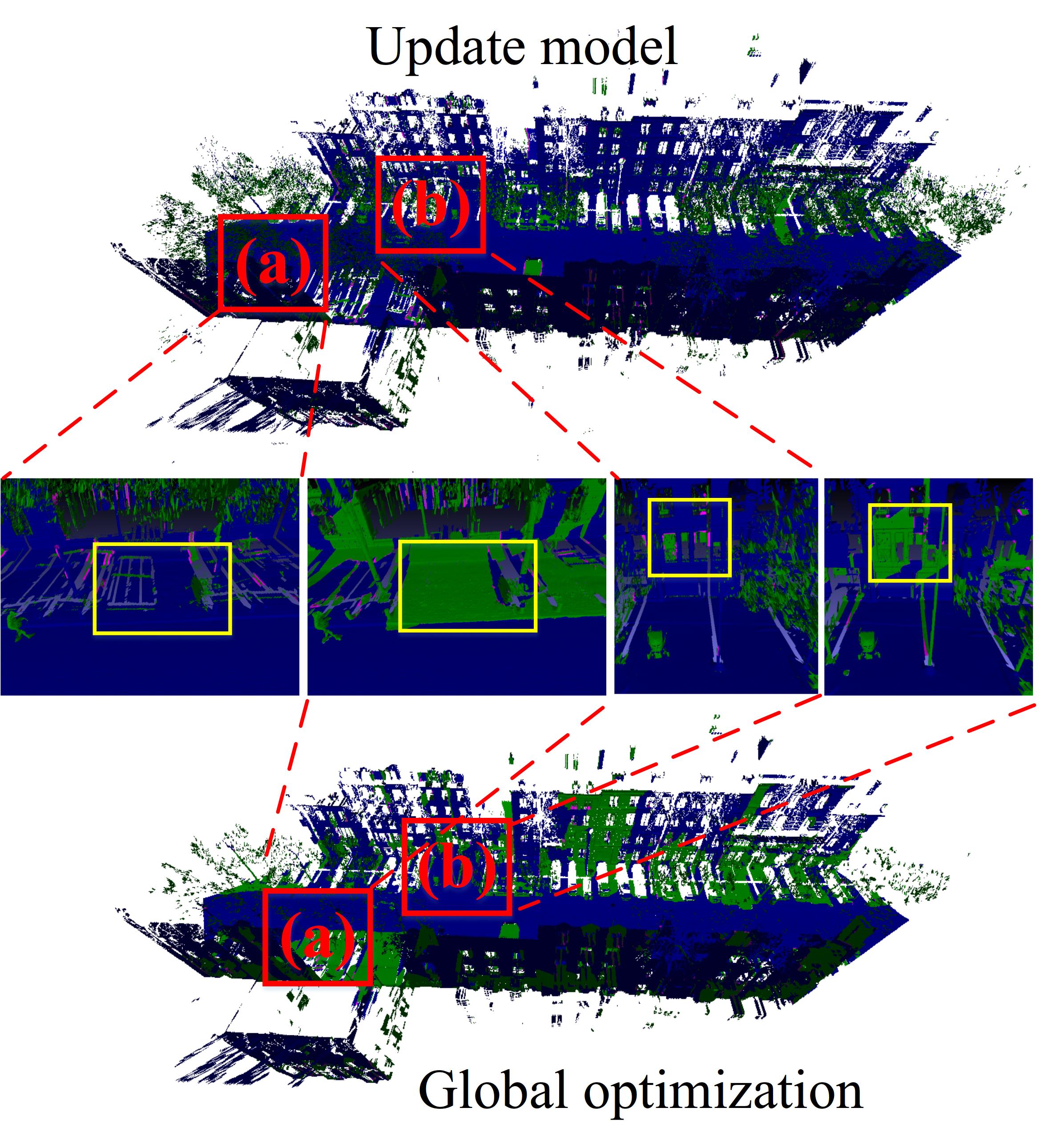}
 	\caption{Compare the result with update model, in area(a), our method obtain a result more complete, (b) show triangle from same mesh with less stitch area. Triangles from first mesh are shown in\legendsquare{BlueViolet}, triangles from the second one are shown in\legendsquare{OliveGreen}. The triangle at stitch line is show in\legendsquare{magenta}.}
 	\label{Fig.compare}
 \end{figure}

For Stéréopolis dataset, it has a higher resolution, means has more triangles, the update model is also compared. In the experiment, points in leaf area are remained. For only two meshes, because the input order is fixed, it will not influence much the filling parts.
As shown in \Cref{Fig.compare}, compare to the update model, the optimization method can obtain more complete result, small area missing because occlusion as shown in (a), and less stitch area as shown in (b).

\section{CONCLUSION AND FUTURE WORK}
In the paper, we propose an automatic pipeline for mesh map sustainable update. Using time series analysis to detect unsustainable changes, seam line is optimized using QPBO based method, all the remain sustainable parts are stitched to obtain a complete mesh. Our method can improve the sustainability and novelty of the map. 

In the method, only the meshes are used, no semantic information from images. In the scene, some unsustainable objects things, like cars, which are parking for a long time, in these cases, they can not be removed in the map.
% add in future work is better
In this way our method can even handle data acquired simultaneously by multiple platforms at the same scene.
In the current experiment, all the meshes are form LiDAR point cloud, because our inputs are meshes with sensor information, image sensors also can produce 3D model reconstruction result with lower expense. We can combine the mobile mapping system(MMS) with a stereo agent to improve the novelty with lower expense.
Mesh processing depends on the 3D reconstruction method, in the current method, only a 2D range sensor is used to reconstruct the scene. Use 3D range sensors or multi sensors to reconstruct the scene to have a better mesh map.
For 3D visualization, to consider the texture stitching is also an interesting topic.

\bibliographystyle{IEEEtran}
\bibliography{irosMesh.bib}

\end{document}